\pdfoutput=1

\documentclass[11pt]{article}

\usepackage[final]{coling}

\usepackage{times}
\usepackage{latexsym}

\usepackage[T1]{fontenc}

\usepackage[utf8]{inputenc}

\usepackage{microtype}

\usepackage{inconsolata}

\usepackage{graphicx}

\usepackage[tikz]{mdframed}
\usetikzlibrary{shadows}

\usepackage{subfigure}
\usepackage{booktabs}
\usepackage{amsmath}
\usepackage{multirow}
\usepackage{arydshln}
\usepackage{pifont}
\usepackage{colortbl}
\usepackage{xcolor}
\definecolor{customgraycolor}{gray}{0.8}

\usepackage{enumitem} 
\usepackage{color}

\title{Self-Evolution Knowledge Distillation for LLM-based Machine Translation}

\author{Yuncheng~Song$^{\heartsuit}$,
\ Liang Ding$^{\Re}$,
\ Changtong Zan$^{\spadesuit}$,
\ Shujian Huang$^{\heartsuit}$\thanks{Corresponding author} \\
\ $^{\heartsuit}$Nanjing University
\ $^{\Re}$The University of Sydney \\
\ $^{\spadesuit}$China University of Petroleum (East China)
\\
\includegraphics[scale=0.15]{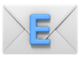} \texttt{songyuncheng@smail.nju.edu.cn}, \texttt{liangding.liam@gmail.com}
\\
\texttt{zanct@s.upc.edu.cn}, \texttt{huangsj@nju.edu.cn}
}

\begin{document}
\maketitle
\begin{abstract}

Knowledge distillation (KD) has shown great promise in transferring knowledge from larger teacher models to smaller student models. However, existing KD strategies for large language models often minimize output distributions between student and teacher models indiscriminately for each token. This overlooks the imbalanced nature of tokens and their varying transfer difficulties. In response, we propose a distillation strategy called Self-Evolution KD. The core of this approach involves dynamically integrating teacher distribution and one-hot distribution of ground truth into the student distribution as prior knowledge, which promotes the distillation process. It adjusts the ratio of prior knowledge based on token learning difficulty, fully leveraging the teacher model’s potential. Experimental results show our method brings an average improvement of approximately 1.4 SacreBLEU points across four translation directions in the WMT22 test sets. Further analysis indicates that the improvement comes from better knowledge transfer from teachers, confirming our hypothesis.

\end{abstract}

\section{Introduction}

Large language models (\citealp[LLMs]{achiam2023gpt,touvron2023llama}) have achieved remarkable success in generating high-quality translations \cite{hendy2023good,Peng2023ChatGPT4MT,jiao2023chatgpt} and other tasks~\cite{kocon2023chatgpt,zhong2023can,lu2024error}. However, previous research indicates that LLM-based translation models (with billion-level parameters) must be several orders of magnitude larger than traditional neural machine translation systems (which typically have millions of parameters) to achieve comparable performance~\cite{garcia2023unreasonable}. This high computational and deployment cost severely hinders the widespread application of LLMs in translation.

In general, a simple and effective technique to reduce high computational footprints is knowledge distillation (KD) ~\cite{hinton2015distilling,kim2016sequence}, which trains a smaller model (aka. student) under the supervision of a larger model (aka. teacher). Recent research on KD for large language models (LLMs) has shown promising results~\cite{gu2024minillm,ko2024distillm,agarwal2024policy,zhong2024revisiting,rao2024exploring}, driven by one key factor: exploration of various divergence losses.

However, several problems are still under-explored. Their training objectives are achieved by indiscriminately minimizing the output distributions between the student and teacher model for each token. In fact, due to the token imbalance nature~\cite{piantadosi2014zipf} and the truth that different tokens contribute differently to the sentence meaning~\cite{chen2020content}, adaptively reweighting the token-level loss would promote the model training, as evidenced by its effectiveness in sequence-to-sequence training \cite{zhang2022conditional,peng2023token}. It motivates us to speculate that indiscriminately adopting the same distillation mode to each token might be sub-optimal. Besides, in human learning patterns, human teachers often provide human students with personal insights (aka prior knowledge) to facilitate student learning. Excellent teachers could adjust the amount of prior knowledge to fully stimulate students' potential. This pattern further supports our hypothesis that the distillation mode should be differentiated based on the student's learning status rather than adopting a uniform strategy. It also hints at the necessity of providing prior knowledge to optimize the distillation strategy and enhance student outcomes.

Therefore, a natural question arises: \textbf{\textit{how to effectively transfer the teacher knowledge based on the student's mastering with the help of prior knowledge?}} It should be a dynamic strategy, which controls the integration of prior knowledge based on the student's training state.

To address it, we propose a simple but effective strategy -- self-evolution knowledge distillation (Self-Evolution KD) for LLMs. It mainly includes two stages: \textbf{\ding{192} Self-Question} and \textbf{\ding{193} Self-Evolution}. In \textbf{Stage 1}, we utilize the Kullback-Leibler (KL) divergence between the student distribution and the target distribution (averaged by the teacher distribution and the one-hot distribution of ground-truth) to quantify the learning difficulty. By comparing this measure with a preset threshold, we assess the student model's learning status at the token level, thereby identifying hard-to-learn and easy-to-learn tokens. This enables us to provide tailored prior knowledge for different tokens in the next stage to enhance the student model's learning. It should be noted that, although \citet{qiu2022better} has demonstrated the potential of prior knowledge in the field of computer vision, treating prior knowledge as input would cause the risk of dimensionality mismatch and significantly increase training costs. To introduce the prior knowledge in a lightweight way, we design a simple strategy --- distribution adjustment. Specifically, in the \textbf{Stage 2}, if the token expresses hard-to-learn property, builds proxy distribution by smoothing the student distribution and target distribution, then used to learn target distribution, thus leading to faster convergence and superior performance. Otherwise, the proxy distribution is the student distribution. By emulating the human teaching mode, assigning prior knowledge to hard-to-learn tokens to fully leverage the target information, while omitting its integration to easy-to-learn tokens, maximizes the potential of the student model.

Empirically, we apply Self-Evolution KD to Llama series models~\cite{touvron2023llama}, with parameter sizes ranging from 7 to 13 billion, and evaluate our proposed method on the WMT22 test sets (En$\leftrightarrow$De and En$\leftrightarrow$Cs). The results show that Self-Evolution KD significantly achieves satisfactory gains over four competitive baselines. Further analysis suggests that Self-Evolution KD is more effective in transferring knowledge from the teacher model to the student model.

\section{Related Work}
\subsection{Language Models for Translation} Before the era of large-scale language models, researchers had already begun leveraging language models to enhance machine translation tasks. This included using discriminative language models, such as BERT~\cite{kenton2019bert}, to improve representational capabilities~\cite{zhuincorporating,guo2021adaptive}, designing Encoder-Decoder models like BART~\cite{lewis-etal-2020-bart} and T5~\cite{raffel2020exploring} to enhance translation quality~\cite{liu2020multilingual}, as well as various subsequent follow-up works to facilitate knowledge transferring~\cite{liu2021complementarity,zan2022complementarity,zan2022bridging,pan-etal-2024-pomp}. 
With the increasing capacity of LLMs, they have already become new standards for various NLP tasks, including machine translation~\cite{jiao2023chatgpt,wang2023document}. One line of work focuses on a comprehensive evaluation of LLMs across various translation scenarios. For example, \citet{zhu2023multilingual,jiao2023chatgpt,hendy2023good} assess the multilingual translation capabilities of LLMs. \citet{hendy2023good,wang2023document,karpinska2023large} evaluates their performance in document-level translation, while \citet{guerreiro2023hallucinations} explores the phenomenon of hallucination. Another line is instruction tuning, such as \citet{zhu2023extrapolating} boosted the translation capability of LLMs by translation data alongside cross-lingual general task data. \citet{xu2023paradigm} proposed a new translation paradigm: initial fine-tuning on monolingual data, followed by a small set of high-quality parallel data. \citet{jiao2023parrot} enhance the translation abilities by leveraging open-source LLMs, human-written translation and feedback data. \citet{zan2024building} enhanced the ability to follow instructions related to translation direction during the instruction tuning process. Different from these approaches, we focus on transferring the translation capabilities from stronger LLM models to weaker LLM models under instruction tuning.

\subsection{Knowledge Distillation for Large Language Models}

The application of KD in LLM falls into two categories: black-box KD which accesses only teacher-generated texts~\cite{chen2024knowledge,hsieh2023distilling,taori2023stanford,peng2023instruction}, and white-box KD which can employ the teacher parameters. Recently, with the increasing accessibility of open-source models, white-box distillation has gained more attention, particularly concerning the role of KL divergence~\cite{zhong2024revisiting,gu2024minillm,wu2024rethinking,ko2024distillm,agarwal2024policy}. Concurrently, several works aim to mitigate the training-inference mismatch problem by leveraging generated text of the student model~\cite{agarwal2024policy,gu2024minillm,ko2024distillm}. Nevertheless, they ignore the efficacy of prior knowledge and indiscriminately handle tokens without differentiation. In this paper, we draw from human learning patterns, dynamically providing prior knowledge based on the learning state of the student model to enhance the distillation. Notably, a concurrent work -- SKEW KLD loss~\cite{gu2024minillm} is also a modified distillation function that integrates the prior knowledge into the student model. However, it still maintains the traditional uniform distillation strategy and limits prior knowledge to the teacher's knowledge. Besides, although \citet{zhong2024revisiting,wang2021selective} assign varied distillation modes by token category, they fix the classification ratio and ignore the effectiveness of prior knowledge.

\subsection{Self-Evolution Learning}
Self-evolution learning is a novel and effective method to exploit the knowledge from data, it is designed to regularize the model training by dynamically learning under-explored tokens. For example, \citet{zhong2023self} dynamically selected hard-to-learn tokens, then encourages the model to learn smoothed distribution which considers precise reference labels and easily digestible distribution generated by the model itself, thereby improving the training efficiency and scalability (up to 6 billion in their follow-up technical report~\cite{zhong2022toward}). \citet{peng2023token} employed a similar strategy and verified the effectiveness of this learning on typical sequence-to-sequence learning tasks, e.g., machine translation, summarization and grammatical error correction tasks. Moreover, \citet{zheng2023self} introduced self-evolution learning to construct more adaptive and model-friendly pseudo samples to strengthen the mix-up-based text classification model. In this work, we focus on applying this learning strategy to distil the translation-tailored LLMs.

\section{Self-Evolution Knowledge Distillation}

\subsection{Preliminary}

We provide some preliminary information about KD for LLMs. It typically employs a pre-trained and fixed teacher model to transfer knowledge into the parameterized student model by providing soft labels from the teacher's output. Given source and ground-truth sequence pair ($s$, $t$) from a fixed dataset ($S$, $T$), KD could be formulated as an optimization problem aimed at minimizing the Kullback-Leibler (KL) divergence between the token-level distributions of the student $\mathbf{q}$ and teacher $\mathbf{p}$ models:

\begin{equation}
\small
\mathcal{L}_{kl}(\mathbf{p}||\mathbf{q})  =\frac{1}{N} \sum_{i=1}^{N} \mathbf{p}(t_i|t_{<i}, s) log \frac{\mathbf{p}(t_i|t_{<i}, s)}{\mathbf{q} (t_i|t_{<i}, s)},
\end{equation}
where $N$ is the length of ground-truth sequence $t$=\{$t_1, ..., t_n$\}.

Furthermore, during the KD process, the student model also requires training under the ground-truth sequence $t$. The corresponding training objective could be calculated:

\begin{equation}
\mathcal{L}_{sft} = \frac{1}{N} \sum_{i=1}^{N} (-log \mathbf{q} (t_i|t_{<i},s)).
\end{equation}

Finally, the overall loss function of KD is a linear interpolation between the supervised fine-tuning (SFT) loss and the KL loss:
\begin{equation}
\mathcal{L} = (1 - \lambda) \mathcal{L}_{sft} + \lambda \mathcal{L}_{kl}(\mathbf{p}||\mathbf{q}),
\label{eq:kd_finally_loss}
\end{equation}
where the parameter $\lambda$ serves as a weight to control the influence of each loss.

\subsection{Self-Evolution Knowledge Distillation}

\begin{figure*}
    \centering
    \includegraphics[width=1\textwidth]{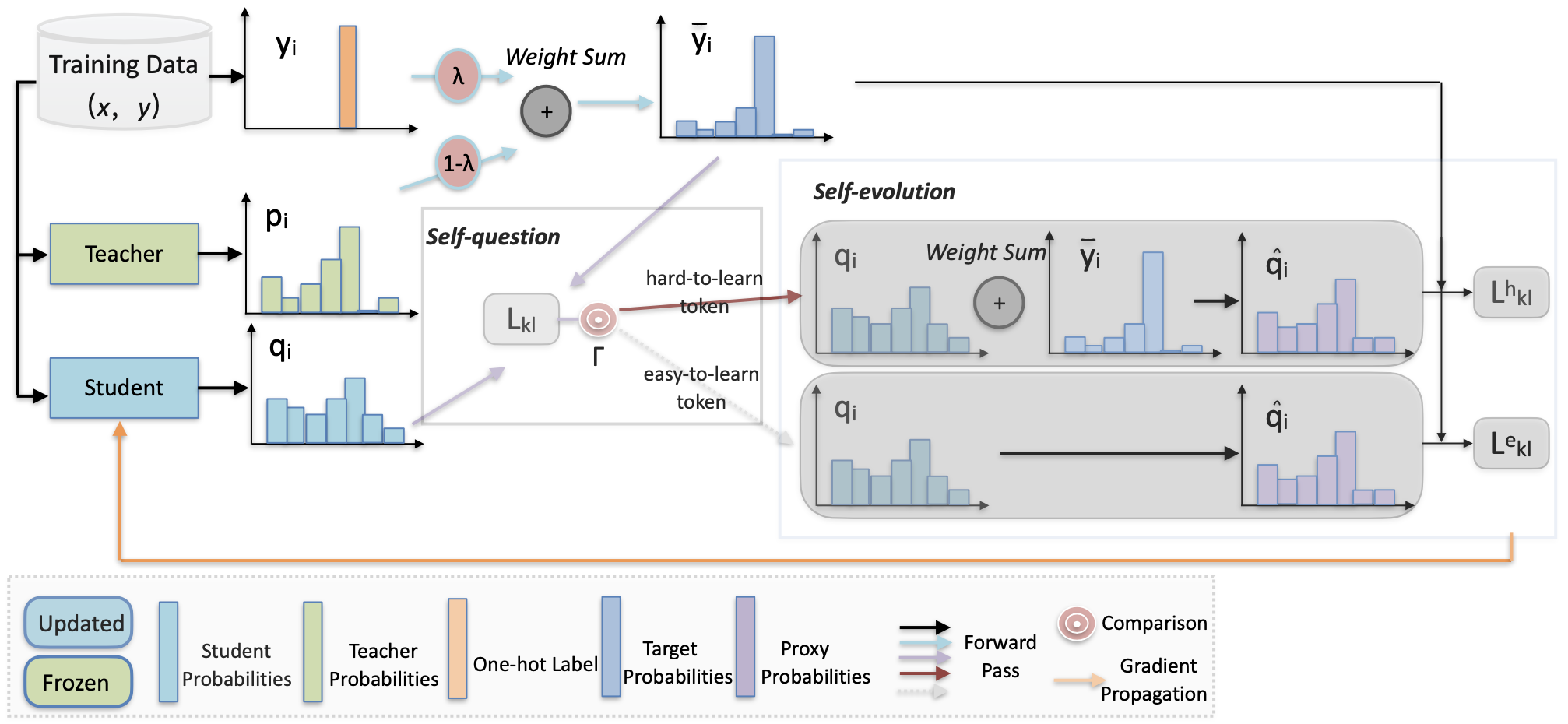}
    \caption{\textbf{Overall framework of our Self-Evolution KD}. It mainly contains two stages: \ding{172} \textit{self-question}: calculating the learning difficulty by the KL divergence between the student distribution and target distribution, and dividing tokens into different categories.
    \textit{comparison} means comparing the learning difficulty with the preset threshold $\Gamma$; \ding{173} \textit{self-evolution}: building proxy distribution for different tokens by smoothing the target and student distributions. \textbf{\textit{Updated}} represents that the parameter needs to be updated, while \textbf{\textit{Frozen}} means not.}
    \label{fig:evokl}
\end{figure*}

Here we introduce the proposed self-evolution knowledge distillation {(Self-Evolution KD)} in detail. As illustrated in Figure \ref{fig:evokl}, it primarily emulates the human teaching mode and comprises two stages: evaluates the student's learning status to identify its hard and easy parts (Stage \textit{1}), offers varied proportions of teacher knowledge for different parts to assist the student learning (Stage \textit{2}).

\textbf{Stage \textit{1} Self-Question Stage} Due to the imbalanced nature of token properties~\cite{piantadosi2014zipf}, we recommend evaluating the student's learning status at the token level. Therefore, the goal of this stage is to classify tokens as either hard-to-learn or easy-to-learn. However, \textit{how to categorize these tokens?} Inspired by previous findings that reveal the dynamic training difficulty of tokens during the training process~\cite{peng2023token}, we leverage the model itself to divide tokens wisely.

Specifically, we first calculate the learning difficulty for each ground-truth token $t_i$, denoted as \{$d_1$, ..., $d_T$\}:

\begin{equation}
\mathbf{\tilde{y}}_{i} = (1 - \lambda) \mathbf{y}_{i} + \lambda \mathbf{p}_{i}
\end{equation}
\begin{equation}
\mathbf{d}_i = \mathcal{L}_{kl}(\mathbf{\tilde{y}}_{i}||\mathbf{q}_i),
\end{equation}
where $\mathbf{y}_i$ is the one-hot distribution of ground-truth token $t_i$ at position $i$. Notably, previous works primarily assesses the student's learning status by measuring the discrepancy between the student and teacher distributions~\cite{zhong2024revisiting}. However, as indicated in Eq. \ref{eq:kd_finally_loss}, the student's learning status should be determined by both the teacher model and ground-truth sequence, since knowledge distillation inherently involves multiple objectives. Therefore, we combine the distributions of the ground-truth sequence and teacher model linearly to derive the target distribution $\mathbf{\tilde{y}}_{i}$, and calculate the divergence gap between it and the student distribution to represent the learning difficulty $d_i$ of the student model.

Then, we preset a threshold $\Gamma$, and select tokens which corresponding KL divergence exceed $\Gamma$ as hard-to-learn tokens, $i.e.$, $\mathcal{T}_h = \{t_i | d_i > \Gamma\}$ where $i \in \{1, ..., N\}$, and the others are easy-to-learn tokens.

\textbf{Stage \textit{2} Self-Evolution Stage} After identifying different types of tokens, our primary focus shifts to leveraging the teacher information to enhance the learning of the student model. Although previous work~\cite{qiu2022better} employs the teacher's hidden states as input improve the distillation, there exists a dimensionality mismatch problem due to the model size discrepancy between the student and teacher, while significantly increases training time. Therefore, \textit{how to integrate teacher information in a lightweight way to enhance the distillation?} It is important to note that the ``teacher'' information here not only comprises the knowledge from the teacher model but also contains the ground-truth information, as discussed in Stage \textit{1}. To address this problem, we propose to promote token distillation by a simple method -- distribution adjustment.

Specifically, we introduce a parameter $\beta$ to mix the student distribution $\mathbf{q}_i$ and target distribution $\mathbf{\tilde{y}}_{i}$ to obtain proxy distribution $\hat{\mathbf{q}}$. Then, using proxy distribution to match target distribution:

\begin{equation}
\hat{\mathbf{q}}_{i} = \beta \mathbf{q}_i + (1-\beta) \mathbf{\tilde{y}}_{i}
\label{exposure_formula}
\end{equation}
\begin{equation}
\mathcal{L}_{kl}^{h_i} = \mathcal{L}_{kl}(\mathbf{\tilde{y}}_{i}||\hat{\mathbf{q}}_{i}).
\end{equation}

By adjusting the information on the distribution, we avoid the dimensionality mismatch problem caused by integrating hidden states, while incurring almost no extra training cost. Furthermore, since the student distribution owns a partial target distribution, it empirically leads to faster convergence and superior performance \cite{ko2024distillm}.

As for the easy-to-learn tokens, given their ability to effectively capture the ``teacher'' information, no additional modifications are necessary. Their corresponding optimization objective is as follows: 
\begin{equation}
\mathcal{L}_{kl}^{e_i} = \mathcal{L}_{kl}(\mathbf{\tilde{y}}_{i}||\mathbf{q}_i).
\end{equation}

\textbf{Overall Optimization}
Finally, we combine the losses of the hard-to-learn tokens and the other tokens. The overall optimization objective is formulated as:
\begin{equation}
\mathcal{L} = \frac{1}{N} (\sum\limits_{i \in \mathcal{T}_e} \mathcal{L}_{kl}^{e_i} + \sum\limits_{j \in \mathcal{T}_h} \mathcal{L}_{kl}^{h_j}),
\end{equation}
where $\mathcal{T}_e$ and $\mathcal{T}_h$ represents the easy-to-learn token set and hard-to-learn token set, respectively.

\section{Experimental Setup}

\subsection{Training Data}

For training data, we use a small yet high-quality parallel dataset\footnote{ https://github.com/fe1ixxu/ALMA} following \citet{xu2023paradigm}. It contains 14k and 12k parallel sentence pairs on English-German (EN-DE) and English-Czech (EN-CS) tasks, respectively. Then, we formatted them into translation instructions in four language directions: En$\to$De, De$\to$En, En$\to$Cs and Cs$\to$En, according to the translation prompt, which resulted in 52K multilingual training sets. Following \citet{jiao2023parrot}, our translation instructions include a fixed preface for all tasks, followed by ``\#\#\# Instruction:'' to describe the translation task, ``\#\#\# Input:'' for presenting the source sentence, and a ``\#\#\# Response:'' with the target sentence to be generated.

\subsection{Model Training}

Our experiments are conducted based on the \textit{Llama-factory}\footnote{https://github.com/hiyouga/LLaMA-Factory} codebase with the Llama family models~\cite{touvron2023llama}. We use supervised fine-tuned (SFT) Llama1-13B trained on the train data as the teacher model, and regard the Llama1-7B as the student model. We fine-tuned all models for 3 epochs with a batch size of 128, while keeping a maximum text length of 512. Besides, we set the learning rate and warmup\_ratio as $2e-5$ and $0.03$, respectively. The experimental parameters and train data remain consistent in both the SFT and KD process. However, while evaluating the performance on the final checkpoints in knowledge distillation, we employ a validation dataset to select the best checkpoint during SFT, as the model easily overfits in this process. The validation dataset is composed of WMT21 En$\to$De and Cs$\to$En test data~\cite{akhbardeh2021findings}. All experiments are conducted on NVIDIA 8*A800 (80GB) GPUs and utilize DeepSpeed ZeRO\footnote{https://github.com/microsoft/DeepSpeed} Stage~3 for efficient model parallelism.

\subsection{Evaluation}
\paragraph{Test Data}
We evaluated the translation performance on the widely used WMT22 test datasets. It is the test sets from the WMT 2022 competition~\cite{kocmi2022findings}\footnote{https://www.statmt.org/wmt22/translation-task.html}, which consists of diverse domains such as news, social, e-commerce, and conversational. The number of sentence pairs for De$\to$En, En$\to$De, En$\to$Cs and Cs$\to$En is 1984, 2037, 2037 and 1448, respectively.

\paragraph{Metrics}
For automatic evaluations, we use SacreBLEU \cite{post2018call}\footnote{https://github.com/mjpost/SacreBleu} and the COMET score~\cite{rei2020comet}\footnote{https://github.com/Unbabel/COMET} with \textit{Unbabel/wmt22-comet-da}. Specifically, SacreBLEU primarily calculates n-gram similarity to measure the surface lexical matching, while COMET relies on cross-lingual pre-trained models to obtain human-like semantic matching.

\paragraph{Baselines}

We consider two traditional knowledge distillation (KD) baselines in our main experiment: 
\begin{itemize}[itemsep=-1mm, topsep=0mm, leftmargin=*]
  \item \textsc{\bf Forward KD}~\cite{hinton2015distilling}: is defined in Eq. \ref{eq:kd_finally_loss};
  \item \textsc{\bf Reverse KD}~\cite{agarwal2024policy}: $(1 - \lambda) \mathcal{L}_{sft} + \lambda \mathcal{L}_{kl}(\mathbf{q}||\mathbf{p})$, swaps the roles of the teacher and student distributions compared to Forward KD.
\end{itemize}

In addition, we dynamically divide tokens into two groups: \textit{easy-to-learn} and \textit{hard-to-learn}, and our method assumes that different tokens require different distillation modes. To better observe the impact of these dynamic changes, we introduce two extreme comparison baselines:
\begin{itemize}[itemsep=-1mm, topsep=0mm, leftmargin=*]
  \item \textsc{\bf NoEvo KD}: $\mathcal{L}_{kl}(\mathbf{\tilde{y}}|| \mathbf{q})$, treat all tokens as easy-to-learn;
  \item \textsc{\bf SKEW KD}: $\mathcal{L}_{kl}(\mathbf{\tilde{y}}||\beta \mathbf{q} + (1-\beta) \mathbf{\tilde{y}})$, regard all tokens as hard-to-learn tokens, which is similar to SKEW KLD~\cite{ko2024distillm}. The default value of $\beta$ is 0.5, which is consistent with our \textsc{\bf Self-Evolution KD} approach.
\end{itemize}

For reference, we also report the performances of Llama-13B and Llama-7B models after SFT as the upper and lower bounds. Additionally, for two traditional baselines, we closely follow \citet{gu2024minillm}, integrating the SFT loss and KL loss with a mixture ratio $\lambda$=0.5. For our proposed baselines and our method, we combine the teacher distribution and the one-hot distribution of ground-truth at a 0.5 ratio to form the target distribution, omitting the SFT loss since the target distribution already encapsulates the ground-truth information. We compute the loss exclusively on the ground-truth sequence. All models use the beam search strategy \cite{vaswani2017attention,freitag2017beam} during inference. Due to the high computational cost and potential out-of-memory (OOM) issues associated with beam search, we set the beam size to 1 as default.

\section{Experimental Results}

\begin{table*}[t]
\small
\tabcolsep 3pt
\centering
\begin{tabular}{lcccccccccc}
    \toprule
       & \multicolumn{2}{c}{\textbf{En $\to$ De}} & \multicolumn{2}{c}{\textbf{De $\to$ En}} & \multicolumn{2}{c}{\textbf{En $\to$ Cs}} & \multicolumn{2}{c}{\textbf{Cs $\to$ En}} & \multicolumn{2}{c}{\textbf{\underline{Average}}} \\
       & \textbf{BLEU} & \textbf{COMET} & \textbf{BLEU} & \textbf{COMET} & \textbf{BLEU} & \textbf{COMET} & \textbf{BLEU} & \textbf{COMET} & \textbf{\underline{BLEU}} & \textbf{\underline{COMET}} \\
    \midrule
      & \multicolumn{10}{c}{\textbf{Test}: \textit{WMT22 Test sets}} \\
      Teacher  & 26.69 & 82.77 & 28.47 & 83.43 & 20.84 & 81.37 & 37.9 & 83.67 &  \underline{28.48}  &  \underline{82.81} \\
    \hline  \addlinespace[2pt]
      Student  & 25.14 & 81.08 & 27.53 & 82.92 & 19.18 & 78.11 & 36.92 & 83.10 &  \underline{27.19}  &  \underline{81.30} \\
      Forward KD  & 25.29 & 81.84 & 27.77 & 83.09 & 20.39 & 80.60 & 36.11 & 83.27 &  \underline{27.39}  &  \underline{82.20} \\
      Reverse KD  & 25.37 & 81.77 & 27.41 & 83.03 & 20.5 & 79.83 & 35.34 & 83.22 &  \underline{27.16}  &  \underline{81.96} \\
      NoEvo KD  & 25.51 & 81.70 & 27.85 & 83.02 & 20.56 & 80.78 & 36.21 & 83.30 &  \underline{27.53}  &  \underline{82.20} \\
      SKEW KD  & 26.1 & 81.74 & 28.09 & 83.20 & 21.22 & 80.45 & 37.77 & 83.53 &  \underline{28.29}  &  \underline{82.23} \\ \hline
    \multicolumn{11}{l}{\textit{\textcolor{lightgray!99}{Ours}}} \\
      Self-Evolution KD &\textbf{26.73}$^{\dagger}$& \textbf{82.05}&\textbf{28.62}$^\dagger$ &\textbf{83.42}&\textbf{21.54}$^\dagger$ &\textbf{80.71} &\textbf{38.44}$^\dagger$ &\textbf{83.74} &\textbf{\underline{28.83}}&\textbf{\underline{82.48}} \\
      \rowcolor{customgraycolor}
      ~~~~$\Delta$ & \textit{+1.44} & \textit{+0.21} & \textit{+0.85} & \textit{+0.33} & \textit{+1.15} & \textit{+0.11} & \textit{+2.33} & \textit{+0.47} & \textit{\underline{$+$1.44}}& \textit{\underline{$+$0.28}} \\
    \bottomrule
\end{tabular}
\caption{\textbf{Comparison results of our \textsc{\bf Self-Evolution KD} against baselines} on different translation tasks, where ``$\Delta$'' indicates the improvement against \textsc{\bf Forward KD}, and ``$\dagger$''indicates statistically significant difference (p$<$0.05). \textsc{\bf Student} and \textsc{\bf Teacher} represent the Llama-7b and Llama-13b after SFT.}  
\label{tab:main_kl_results}
\end{table*}

\subsection{Main Results}

We report the comparison of our \textsc{\bf Self-Evolution KD} and other competitive baseline distillation methods in Table~\ref{tab:main_kl_results}, in terms of translation performance (SacreBLEU and COMET scores). We have the following observations:

\paragraph{Larger models produce better translations.} We observe that the translation ability of the Llama model improves with increased model capacity. These gaps between teachers and students exist across all language pairs, which means there is much room for our KD methods.

\paragraph{Forward KD is effective} Table \ref{tab:main_kl_results} shows that Forward KD could achieve an average gain of 0.2 SacreBLEU points and 0.9 COMET score over SFT, highlighting its effectiveness. However, when compared to the performance gap between the teacher and student models (about 1.3 SacreBLEU points and 1.5 COMET score on average), the gains are relatively modest. This limitation suggests that Forward KD does not fully leverage the potential of the teacher model.

Furthermore, recent methods \cite{gu2024minillm,kim2024promptkd} argue that Reverse KD is more suitable for large language models than Forward KD. Nevertheless, our findings indicate that pure Forward or Reverse KD yields similar performance without significant differences. Consequently, in this paper, we emphasize the substantial benefits derived from dynamically integrating prior knowledge for each token.

\paragraph{Self-Evolution KD improves distillation} Table~\ref{tab:main_kl_results} shows that:
\begin{itemize}
\item Compared to baseline Forward KD, Self-Evolution KD achieves significant improvements (average: +1.44 SacreBLEU points / +0.28 COMET scores). It shows a maximum improvement of approximately 2.33 SacreBLEU points and about 0.47 COMET scores in the Cs$\to$En test set. Besides, Self-Evolution KD achieves a performance comparable to that of the teacher model and even surpasses it on the SacreBLEU metric. These substantial gains underscore the efficiency of our approach.
\item As an adaptive strategy, Self-Evolution KD dynamically allocates different strategies based on the learning status of each token. It surpasses both static strategies, as shown in Table \ref{tab:main_kl_results}, corroborating our hypothesis that indiscriminately distilling each token is a sub-optimal strategy. Adopting varied distillation modes on the basis of the student's learning status would better align the distillation curve of the student model, thereby fully unleashing its potential. Moreover, with the integration of prior knowledge, SKEW KD also shows significant improvement over Forward KD, with average gains of 0.9 SacreBLEU points.
\end{itemize}

\subsection{Ablation Study}
\subsubsection{Effect of Token Selection} 

\begin{figure*}
    \centering
    \subfigure[Values of $\Gamma$]{
        \label{fig:lossthres}
        \includegraphics[width=0.31\textwidth]{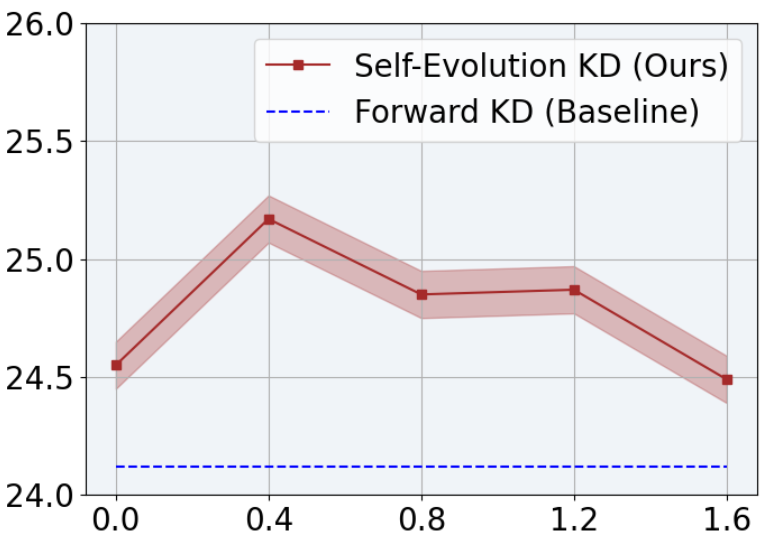}
    }
    \subfigure[Values of $K (\%)$]{
        \label{fig:selectratio}
	\includegraphics[width=0.31\textwidth]{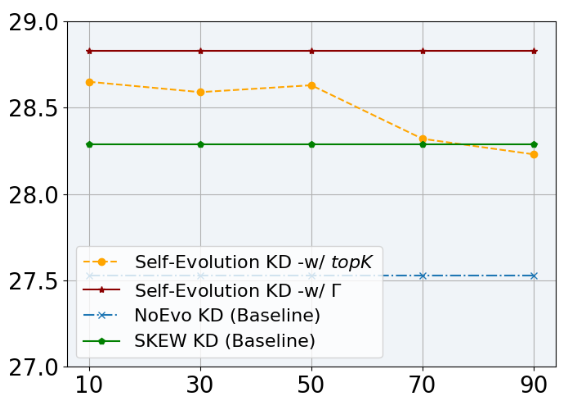}
    }
    \subfigure[Values of $\beta$]{
        \label{fig:exposureratio}
	\includegraphics[width=0.31\textwidth]{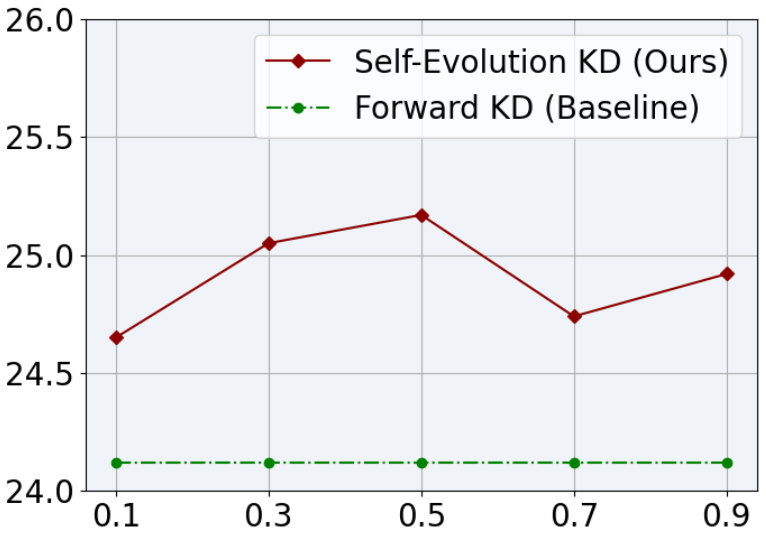}
    }
    \caption{\ref{fig:lossthres} and \ref{fig:selectratio}: \textbf{Effect of $\Gamma$ and percent ($K$)} for selecting hard-to-learn tokens. \ref{fig:exposureratio}: \textbf{Effect of $\beta$} to determine the mixsture proportion of prior knowledge. We report their average SacreBLEU points on the above-mentioned validation dataset in \ref{fig:lossthres} and \ref{fig:exposureratio}. As for \ref{fig:selectratio}, the average SacreBLEU points on WMT22 test sets are reported since we compare different distillation strategies.}
    \label{fig.abalation}
\end{figure*}

As a key point in this paper, how to divide tokens after evaluating the student's learning state is worth considering. Traditional methods typically select top $K$ percent of all tokens as hard-to-learn tokens \cite{zhong2024revisiting}. However, we suspect that this approach is sub-optimal, since it forces the model to choose a fixed proportion of hard-to-learn tokens even in the later stages of distillation. In contrast, dynamically selecting hard-to-learn tokens based on a preset threshold $\Gamma$ would avoid choosing hard-to-learn tokens after all tokens have been sufficiently learned, thereby fully harnessing the potential of the student model. In this section, we delve into the impact of the two strategies. The former is named \textsc{\bf Self-Evolution KD -w/ top$K$}, while the latter is \textsc{\bf Self-Evolution KD -w/ $\Gamma$}.

First, we examine the influence of the hyper-parameter $\Gamma$ within our dynamic strategy. As shown in Fig \ref{fig:lossthres}, we observe that a larger or smaller threshold detrimentally affects the performance of Self-Evolution KD. The former leads to an overabundance of easy-to-learn tokens and fails to adequately focus on tokens that need assistance, while the latter is the opposite. Self-Evolution KD performs best with an optimal value of $\Gamma$ = 0.4, thus we retain it as our default setting.

Second, from Fig \ref{fig:selectratio}, the Self-Evolution \textit{-w/ top$K$} outperforms the NoEvo KD and SKEW KD, it further demonstrates the effectiveness of assigning distinct distillation strategies to different tokens. However, its performance is inferior to the dynamic selection strategy (\textit{-w/ $\Gamma$}), which is consistent with our aforementioned hypothesis.
Takeaway: \textbf{\textit{These observations suggest that selecting the appropriate number of hard-to-learn tokens is crucial, since a larger or smaller number would lead to negative impacts. Besides, a dynamic token selection strategy aligns better with the learning patterns of the student model, thus fully unlocking its potential.}}

\subsubsection{Influence of $\mathbf{\beta}$} 

The factor $\beta$ in Eq. \ref{exposure_formula}, which serves to control the proportion of target distribution integrated to the student distribution, also requires to be investigated. Figure \ref{fig:exposureratio} shows the results of varied $\beta$ ranging from 0.1 to 0.9. As observed, the model performs optimally with $\beta$ = 0.5, thus we adopt it as default setting in our experiments.

\begin{figure}
    \centering
    \includegraphics[width=0.9\columnwidth]{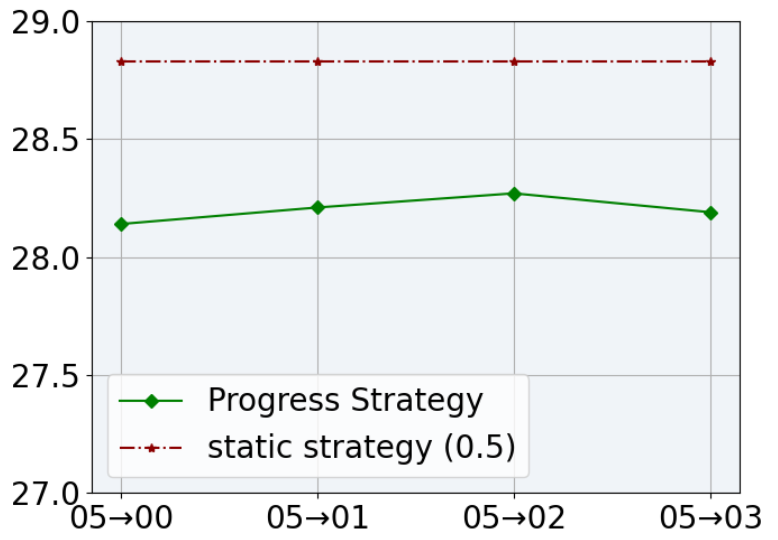}
    \caption{\textbf{Comparison of static strategy and progressive strategy for factor $\beta$}. ``0.5$\to$0.0'' means the $\beta_b$ is 0.5 and the $\beta_e$ is 0.0. \textit{static strategy (0.5)} indicates the results of Self-Evolution KD ($\beta$ = 0.5).}
    \label{fig.glancing}
\end{figure}

Besides, we also employ the progressive strategy outlined in \citet{qian2020glancing}, dynamically adjusting the integration ratio of the target distribution. Specifically, we set an initial ratio ($\beta_b$) and linearly decrease it to a predefined final value ($\beta_e$) throughout the training process. As seen in Figure \ref{fig.glancing}, the dynamic strategy fails to bring substantial gain, despite its potential advantages. Takeaway: \textbf{\textit{it proves that indiscriminately distill each token, even with dynamic adjustments to prior knowledge, still constrains the model performance, leading to sub-optimal outcomes. }}

\subsection{Further Analysis}

\subsubsection{Does Self-Evolution KD Transfer The Teacher's Knowledge Better?}

\begin{figure}
    \centering
\includegraphics[width=0.94\columnwidth]{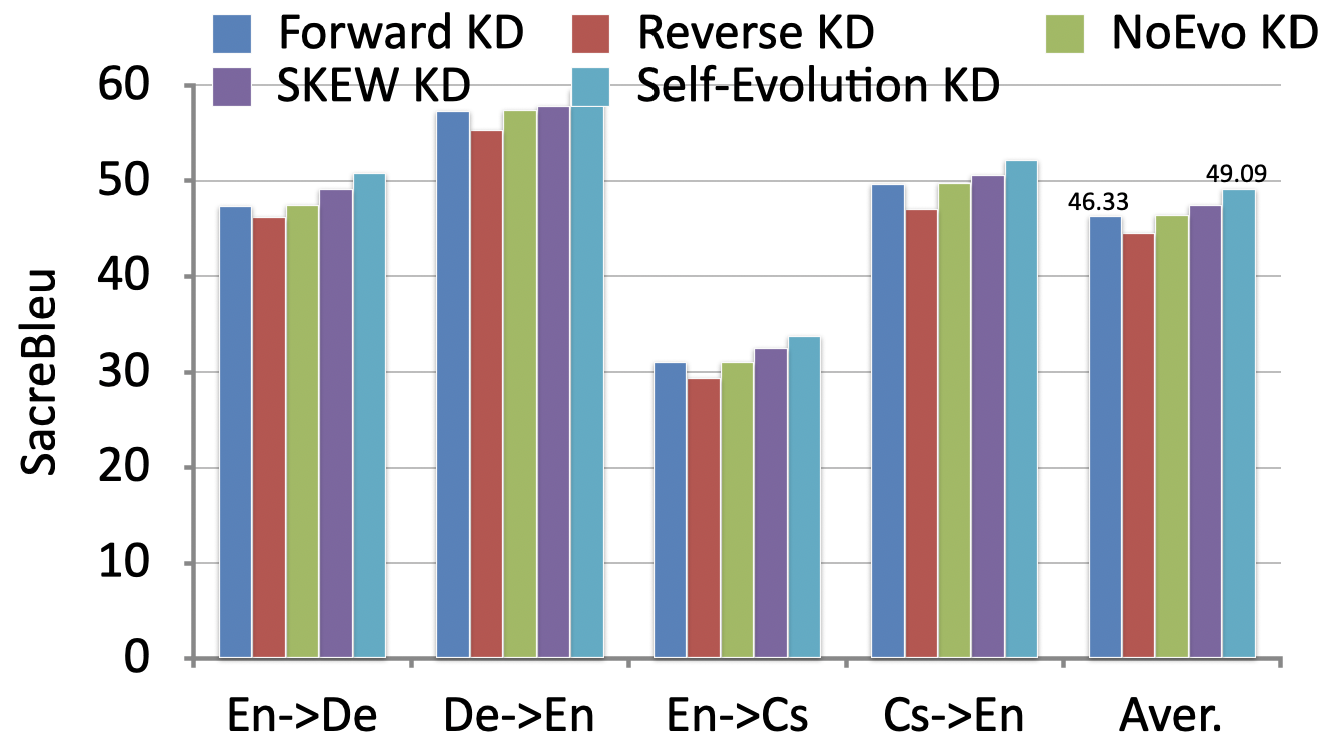}
    \caption{\textbf{Comparison of teacher’s knowledge transfer across different distillation strategies.}}
    \label{fig.transfer}
\end{figure}

The core of Knowledge Distillation (KD) is to transfer the distilled knowledge from a well-performing but cumbersome teacher model to a compact and lightweight student model, thus we analyze the effectiveness of knowledge transfer across various distillation strategies in this part. We regard the generation text of teacher model on the WMT22 En$\leftrightarrow$De, En$\leftrightarrow$Cs test sets as the ``reference'', calculate the SacreBLEU scores between the ``reference'' and generated text of various distillation strategies. As illustrated in Figure \ref{fig.transfer}, SKEW KD is superior to Forward KD in terms of similarity to the teacher model's generated text. Notably, Self-Evolution KD achieves the best performance, with an average improvement of 2.8 gains across the four language pairs. Takeaway: \textbf{\textit{These findings indicate that the introduction of prior knowledge enables the student model to better capture the target information and enhances the efficiency of the teacher's knowledge transfer. Additionally, a dynamic strategy that integrates prior knowledge based on the student's learning status further enhances its potential and deepens its understanding of the teacher's knowledge.}}

\subsubsection{Comparison of Prior Knowledge}

Given that knowledge distillation inherently involves multi-objective learning, we suspect that only considering the teacher knowledge as prior knowledge would significantly reduce the KL divergence loss and mislead the student model to emphasize the SFT loss, thus potentially curtailing the benefits derivable from teacher knowledge. Following \citet{ko2024distillm}, we redefine the distillation objective as (SKEW KD (teacher)):
\begin{equation}
\mathcal{L} = (1 - \lambda) \mathcal{L}_{sft} + \lambda \mathcal{L}_{kl} (p || \beta q_\theta + (1-\beta) p),
\end{equation}
where $\beta$ = 0.9\cite{ko2024distillm}. 

\begin{figure}
    \centering
    \includegraphics[width=0.9\columnwidth]{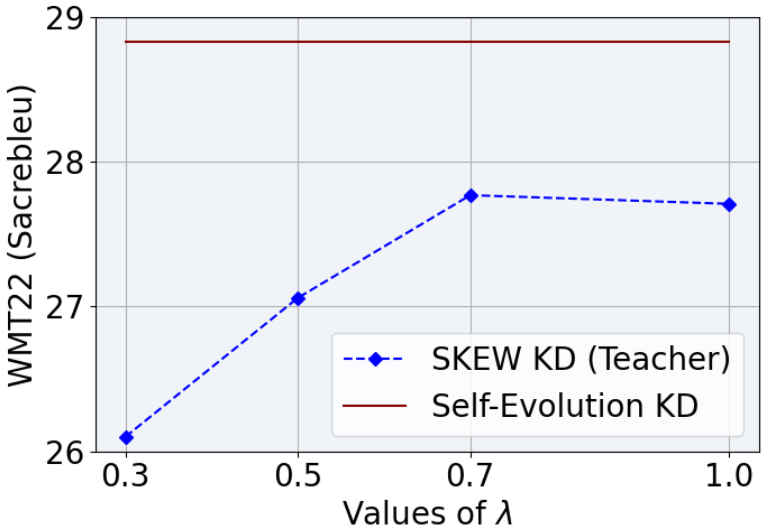}
    \caption{\textbf{Effect of the loss weight of the SKEW KD (Teacher).} We only report the Self-Evolution KD for reference.}
    \label{fig.skew_loss_weight}
\end{figure}

As shown in Figure \ref{fig.skew_loss_weight}, we observe the results when setting $\lambda$ to 0.3, 0.5, 0.7 and 1.0, respectively. While the parameter $\lambda$ is set to 1.0, as \citet{ko2024distillm}, the result proves that ignoring the ground-truth information leads to a significant decline in performance. Besides, despite adjusting the loss weights to balance the SFT loss and the KL divergence loss, the performance of SKEW KD (teacher) still significantly lags behind that of Self-Evolution KD. Takeaway: \textbf{\textit{This finding is consistent with our conjecture that modifying the distillation optimization function by only integrating the teacher's knowledge disrupts the original cooperative relationship between multiple objectives, thus weakening the student's performance.}}

\subsubsection{Larger Teacher Model}

\begin{table}[t]
\small
\begin{center}
\tabcolsep 1.5pt
\begin{tabular}{lccccc}
    \toprule
      WMT22 & En$\to$De & De$\to$En & En$\to$Cs & Cs$\to$En & Avg.  \\
    \midrule
    Teacher & 26.81 & 27.53 & 20.64 & 38.17 & \underline{28.29} \\
    \midrule
    Forward KD & 25.36 & 27.17 & 19.91 & 35.42 & \underline{26.96} \\
    Reverse KD & 24.92 & 26.62 & 19.81 & 34.40 & \underline{26.44} \\
    NoEvo KD & 25.32 & 27.13 & 19.77 & 35.92 & \underline{27.04} \\
    SKEW KD & 25.97 & \textbf{28.14} & 20.91 & 37.24 & \underline{28.07} \\
      \rowcolor{customgraycolor}
      Self-Evolution KD & \textbf{26.73} & \textbf{28.14} & \textbf{21.65} & \textbf{38.07} & \underline{\textbf{28.65}} \\
    \bottomrule
\end{tabular}
\caption{\textbf{Comparison results for different distillation strategies} between the teacher model Llama-30B after SFT and the student model Llama-7B(\textbf{SacreBLEU}).}  
\label{tab:larger_model_size}
 \end{center}
\end{table}

To assess the effectiveness of knowledge distillation in models with substantial size disparities, we employ Llama-30b models after SFT as teachers, distilling knowledge into student models with 7 billion parameters. Table \ref{tab:larger_model_size} provides a comprehensive comparison against several strong baselines. As observed, Self-Evolution KD significantly outperforms the baseline (Forward KD), with an average improvement of approximately 1.7 SacreBLEU points, and surpasses other distillation strategies. Notably, the distillation gains from the teacher model with 30 billion parameters surpass those from the teacher model with 13 billion parameters (e.g., 1.7 SacreBleu point V.S. 1.5 SacreBleu point). Takeaway: \textbf{\textit{It clearly indicates that our method consistently achieves superior performance, even in large model size gaps. It also shows the potential to mitigate the adverse effects of distillation caused by large teacher models.}}

\section{Conclusion}
This paper explores the potential of utilizing prior knowledge in LLM knowledge distillation and highlights the limitations of equally distilling each token without differentiation. In particular, we propose a dynamic teaching mode that enhances student distillation by integrating prior knowledge which contains teacher knowledge and ground-truth information. Besides, we adjust the integration ratio based on the student's token-level learning status. Experimental results demonstrate that our approach consistently enhances distillation performance across multiple translation tasks. Furthermore, in-depth analysis indicates that our method effectively transfers the teacher's knowledge.

\section*{Limitation}
Our work has several potential limitations. First, given the limited computational budget, our method has not been validated across extensive model size gaps, such as 65B. Scaling up to larger model sizes will be more convincing. Second, one key factor $\Gamma$ is empirical and preset value, we fix it throughout the training, following previous works~\cite{peng2023token}. It would be sensible and elegant to dynamically determine the threshold according to training difficulty. For example, you can employ an additional network to predict the integration ratio of prior knowledge for each token.

\section*{Acknowledgments}
We would like to thank the anonymous reviewers for their insightful comments. Shujian Huang is the corresponding author. This work is supported by National Science Foundation of China (No. 62376116, 62176120).

\bibliography{custom}

\begin{thebibliography}{56}
\providecommand{\natexlab}[1]{#1}

\bibitem[{Achiam et~al.(2023)Achiam, Adler, Agarwal, Ahmad, Akkaya, Aleman, Almeida, Altenschmidt, Altman, Anadkat et~al.}]{achiam2023gpt}
Josh Achiam, Steven Adler, Sandhini Agarwal, Lama Ahmad, Ilge Akkaya, Florencia~Leoni Aleman, Diogo Almeida, Janko Altenschmidt, Sam Altman, Shyamal Anadkat, et~al. 2023.
\newblock Gpt-4 technical report.
\newblock \emph{arXiv preprint arXiv:2303.08774}.

\bibitem[{Agarwal et~al.(2024)Agarwal, Vieillard, Zhou, Stanczyk, Garea, Geist, and Bachem}]{agarwal2024policy}
Rishabh Agarwal, Nino Vieillard, Yongchao Zhou, Piotr Stanczyk, Sabela~Ramos Garea, Matthieu Geist, and Olivier Bachem. 2024.
\newblock On-policy distillation of language models: Learning from self-generated mistakes.
\newblock In \emph{The Twelfth International Conference on Learning Representations}.

\bibitem[{Akhbardeh et~al.(2021)Akhbardeh, Arkhangorodsky, Biesialska, Bojar, Chatterjee et~al.}]{akhbardeh2021findings}
Farhad Akhbardeh, Arkady Arkhangorodsky, Magdalena Biesialska, Ond{\v{r}}ej Bojar, Rajen Chatterjee, et~al. 2021.
\newblock Findings of the 2021 conference on machine translation (wmt21).
\newblock In \emph{Proceedings of the sixth conference on machine translation}.

\bibitem[{Chen et~al.(2024)Chen, Quan, Chen, Yan, and Zhang}]{chen2024knowledge}
Hongzhan Chen, Xiaojun Quan, Hehong Chen, Ming Yan, and Ji~Zhang. 2024.
\newblock Knowledge distillation for closed-source language models.
\newblock \emph{arXiv preprint arXiv:2401.07013}.

\bibitem[{Chen et~al.(2020)Chen, Wang, Utiyama, and Sumita}]{chen2020content}
Kehai Chen, Rui Wang, Masao Utiyama, and Eiichiro Sumita. 2020.
\newblock Content word aware neural machine translation.
\newblock In \emph{Proceedings of the 58th Annual Meeting of the Association for Computational Linguistics}.

\bibitem[{Freitag and Al-Onaizan(2017)}]{freitag2017beam}
Markus Freitag and Yaser Al-Onaizan. 2017.
\newblock Beam search strategies for neural machine translation.
\newblock \emph{arXiv preprint arXiv:1702.01806}.

\bibitem[{Garcia et~al.(2023)Garcia, Bansal, Cherry, Foster, Krikun, Johnson, and Firat}]{garcia2023unreasonable}
Xavier Garcia, Yamini Bansal, Colin Cherry, George Foster, Maxim Krikun, Melvin Johnson, and Orhan Firat. 2023.
\newblock The unreasonable effectiveness of few-shot learning for machine translation.
\newblock In \emph{International Conference on Machine Learning}, pages 10867--10878. PMLR.

\bibitem[{Gu et~al.(2024)Gu, Dong, Wei, and Huang}]{gu2024minillm}
Yuxian Gu, Li~Dong, Furu Wei, and Minlie Huang. 2024.
\newblock Minillm: Knowledge distillation of large language models.
\newblock In \emph{The Twelfth International Conference on Learning Representations}.

\bibitem[{Guerreiro et~al.(2023)Guerreiro, Alves, Waldendorf, Haddow, Birch, Colombo, and Martins}]{guerreiro2023hallucinations}
Nuno~M Guerreiro, Duarte~M Alves, Jonas Waldendorf, Barry Haddow, Alexandra Birch, Pierre Colombo, and Andr{\'e}~FT Martins. 2023.
\newblock Hallucinations in large multilingual translation models.
\newblock \emph{Transactions of the Association for Computational Linguistics}.

\bibitem[{Guo et~al.(2021)Guo, Zhang, Xu, Chen, and Chen}]{guo2021adaptive}
Junliang Guo, Zhirui Zhang, Linli Xu, Boxing Chen, and Enhong Chen. 2021.
\newblock Adaptive adapters: An efficient way to incorporate bert into neural machine translation.
\newblock \emph{IEEE/ACM Transactions on Audio, Speech, and Language Processing}.

\bibitem[{Hendy et~al.(2023)Hendy, Abdelrehim, Sharaf, Raunak, Gabr, Matsushita, Kim, Afify, and Awadalla}]{hendy2023good}
Amr Hendy, Mohamed Abdelrehim, Amr Sharaf, Vikas Raunak, Mohamed Gabr, Hitokazu Matsushita, Young~Jin Kim, Mohamed Afify, and Hany~Hassan Awadalla. 2023.
\newblock How good are gpt models at machine translation? a comprehensive evaluation.
\newblock \emph{arXiv preprint arXiv:2302.09210}.

\bibitem[{Hinton et~al.(2015)Hinton, Vinyals, and Dean}]{hinton2015distilling}
Geoffrey Hinton, Oriol Vinyals, and Jeff Dean. 2015.
\newblock Distilling the knowledge in a neural network.
\newblock \emph{arXiv preprint arXiv:1503.02531}.

\bibitem[{Hsieh et~al.(2023)Hsieh, Li, Yeh, Nakhost, Fujii, Ratner, Krishna, Lee, and Pfister}]{hsieh2023distilling}
Cheng-Yu Hsieh, Chun-Liang Li, Chih-Kuan Yeh, Hootan Nakhost, Yasuhisa Fujii, Alexander Ratner, Ranjay Krishna, Chen-Yu Lee, and Tomas Pfister. 2023.
\newblock Distilling step-by-step! outperforming larger language models with less training data and smaller model sizes.
\newblock \emph{arXiv preprint arXiv:2305.02301}.

\bibitem[{Jiao et~al.(2023{\natexlab{a}})Jiao, Huang, Wang, He, Liang, Wang, Shi, and Tu}]{jiao2023parrot}
Wenxiang Jiao, Jen-tse Huang, Wenxuan Wang, Zhiwei He, Tian Liang, Xing Wang, Shuming Shi, and Zhaopeng Tu. 2023{\natexlab{a}}.
\newblock Parrot: Translating during chat using large language models tuned with human translation and feedback.
\newblock In \emph{Findings of the Association for Computational Linguistics: EMNLP 2023}.

\bibitem[{Jiao et~al.(2023{\natexlab{b}})Jiao, Wang, Huang, Wang, and Tu}]{jiao2023chatgpt}
Wenxiang Jiao, Wenxuan Wang, Jen-tse Huang, Xing Wang, and Zhaopeng Tu. 2023{\natexlab{b}}.
\newblock Is chatgpt a good translator? a preliminary study.
\newblock \emph{arXiv preprint arXiv:2301.08745}, 1(10).

\bibitem[{Karpinska and Iyyer(2023)}]{karpinska2023large}
Marzena Karpinska and Mohit Iyyer. 2023.
\newblock Large language models effectively leverage document-level context for literary translation, but critical errors persist.
\newblock \emph{arXiv preprint arXiv:2304.03245}.

\bibitem[{Kenton and Toutanova(2019)}]{kenton2019bert}
Jacob Devlin Ming-Wei~Chang Kenton and Lee~Kristina Toutanova. 2019.
\newblock Bert: Pre-training of deep bidirectional transformers for language understanding.
\newblock In \emph{Proceedings of naacL-HLT}. Minneapolis, Minnesota.

\bibitem[{Kim et~al.(2024)Kim, Jang, and Yang}]{kim2024promptkd}
Gyeongman Kim, Doohyuk Jang, and Eunho Yang. 2024.
\newblock Promptkd: Distilling student-friendly knowledge for generative language models via prompt tuning.
\newblock \emph{arXiv preprint arXiv:2402.12842}.

\bibitem[{Kim and Rush(2016)}]{kim2016sequence}
Yoon Kim and Alexander~M Rush. 2016.
\newblock Sequence-level knowledge distillation.
\newblock \emph{arXiv preprint arXiv:1606.07947}.

\bibitem[{Ko et~al.(2024)Ko, Kim, Chen, and Yun}]{ko2024distillm}
Jongwoo Ko, Sungnyun Kim, Tianyi Chen, and Se-Young Yun. 2024.
\newblock Distillm: Towards streamlined distillation for large language models.
\newblock \emph{arXiv preprint arXiv:2402.03898}.

\bibitem[{Kocmi et~al.(2022)Kocmi, Bawden, Bojar, Dvorkovich, Federmann, Fishel et~al.}]{kocmi2022findings}
Tom Kocmi, Rachel Bawden, Ond{\v{r}}ej Bojar, Anton Dvorkovich, Christian Federmann, Mark Fishel, et~al. 2022.
\newblock Findings of the 2022 conference on machine translation (wmt22).
\newblock In \emph{Proceedings of the Seventh Conference on Machine Translation (WMT)}.

\bibitem[{Koco{\'n} et~al.(2023)Koco{\'n}, Cichecki, Kaszyca, Kochanek, Szyd{\l}o, Baran, Bielaniewicz, Gruza, Janz, Kanclerz et~al.}]{kocon2023chatgpt}
Jan Koco{\'n}, Igor Cichecki, Oliwier Kaszyca, Mateusz Kochanek, Dominika Szyd{\l}o, Joanna Baran, Julita Bielaniewicz, Marcin Gruza, Arkadiusz Janz, Kamil Kanclerz, et~al. 2023.
\newblock Chatgpt: Jack of all trades, master of none.
\newblock \emph{Information Fusion}.

\bibitem[{Lewis et~al.(2020)Lewis, Liu, Goyal et~al.}]{lewis-etal-2020-bart}
Mike Lewis, Yinhan Liu, Naman Goyal, et~al. 2020.
\newblock {BART}: Denoising sequence-to-sequence pre-training for natural language generation, translation, and comprehension.
\newblock In \emph{Proceedings of the 58th Annual Meeting of the Association for Computational Linguistics}.

\bibitem[{Liu et~al.(2021)Liu, Wang, Wong, Ding, Chao, Shi, and Tu}]{liu2021complementarity}
Xuebo Liu, Longyue Wang, Derek~F Wong, Liang Ding, Lidia~S Chao, Shuming Shi, and Zhaopeng Tu. 2021.
\newblock On the complementarity between pre-training and back-translation for neural machine translation.
\newblock In \emph{Findings of the Association for Computational Linguistics: EMNLP 2021}.

\bibitem[{Liu et~al.(2020)Liu, Gu, Goyal, Li, Edunov, Ghazvininejad, Lewis, and Zettlemoyer}]{liu2020multilingual}
Yinhan Liu, Jiatao Gu, Naman Goyal, Xian Li, Sergey Edunov, Marjan Ghazvininejad, Mike Lewis, and Luke Zettlemoyer. 2020.
\newblock Multilingual denoising pre-training for neural machine translation.
\newblock \emph{Transactions of the Association for Computational Linguistics}.

\bibitem[{Lu et~al.(2024)Lu, Qiu, Ding, Zhang, Kocmi, and Tao}]{lu2024error}
Qingyu Lu, Baopu Qiu, Liang Ding, Kanjian Zhang, Tom Kocmi, and Dacheng Tao. 2024.
\newblock Error analysis prompting enables human-like translation evaluation in large language models.
\newblock In \emph{Findings of the Association for Computational Linguistics ACL 2024}.

\bibitem[{Pan et~al.(2024)Pan, Tian, Ding, Zheng, Huang, Wen, and Li}]{pan-etal-2024-pomp}
Shilong Pan, Zhiliang Tian, Liang Ding, Haoqi Zheng, Zhen Huang, Zhihua Wen, and Dongsheng Li. 2024.
\newblock {POMP}: Probability-driven meta-graph prompter for {LLM}s in low-resource unsupervised neural machine translation.
\newblock In \emph{Proceedings of the 62nd Annual Meeting of the Association for Computational Linguistics (Volume 1: Long Papers)}.

\bibitem[{Peng et~al.(2023{\natexlab{a}})Peng, Li, He, Galley, and Gao}]{peng2023instruction}
Baolin Peng, Chunyuan Li, Pengcheng He, Michel Galley, and Jianfeng Gao. 2023{\natexlab{a}}.
\newblock Instruction tuning with gpt-4.
\newblock \emph{arXiv preprint arXiv:2304.03277}.

\bibitem[{Peng et~al.(2023{\natexlab{b}})Peng, Ding, Zhong, Ouyang, Rong, Xiong, and Tao}]{peng2023token}
Keqin Peng, Liang Ding, Qihuang Zhong, Yuanxin Ouyang, Wenge Rong, Zhang Xiong, and Dacheng Tao. 2023{\natexlab{b}}.
\newblock Token-level self-evolution training for sequence-to-sequence learning.
\newblock In \emph{Proceedings of the 61st Annual Meeting of the Association for Computational Linguistics (Volume 2: Short Papers)}.

\bibitem[{Peng et~al.(2023{\natexlab{c}})Peng, Ding, Zhong, Shen, Liu, Zhang, Ouyang, and Tao}]{Peng2023ChatGPT4MT}
Keqin Peng, Liang Ding, Qihuang Zhong, Li~Shen, Xuebo Liu, Min Zhang, Yuanxin Ouyang, and Dacheng Tao. 2023{\natexlab{c}}.
\newblock Towards making the most of chatgpt for machine translation.
\newblock In \emph{Findings of EMNLP}.

\bibitem[{Piantadosi(2014)}]{piantadosi2014zipf}
Steven~T Piantadosi. 2014.
\newblock Zipf’s word frequency law in natural language: A critical review and future directions.
\newblock \emph{Psychonomic bulletin \& review}.

\bibitem[{Post(2018)}]{post2018call}
Matt Post. 2018.
\newblock A call for clarity in reporting bleu scores.
\newblock \emph{WMT 2018}.

\bibitem[{Qian et~al.(2020)Qian, Zhou, Bao, Wang, Qiu, Zhang, Yu, and Li}]{qian2020glancing}
Lihua Qian, Hao Zhou, Yu~Bao, Mingxuan Wang, Lin Qiu, Weinan Zhang, Yong Yu, and Lei Li. 2020.
\newblock Glancing transformer for non-autoregressive neural machine translation.
\newblock \emph{arXiv preprint arXiv:2008.07905}.

\bibitem[{Qiu et~al.(2022)Qiu, Ma, Yang, Liu, Hou, Yi, and Ouyang}]{qiu2022better}
Zengyu Qiu, Xinzhu Ma, Kunlin Yang, Chunya Liu, Jun Hou, Shuai Yi, and Wanli Ouyang. 2022.
\newblock Better teacher better student: Dynamic prior knowledge for knowledge distillation.
\newblock \emph{arXiv preprint arXiv:2206.06067}.

\bibitem[{Raffel et~al.(2020)Raffel, Shazeer, Roberts, Lee, Narang, Matena, Zhou, Li, and Liu}]{raffel2020exploring}
Colin Raffel, Noam Shazeer, Adam Roberts, Katherine Lee, Sharan Narang, Michael Matena, Yanqi Zhou, Wei Li, and Peter~J Liu. 2020.
\newblock Exploring the limits of transfer learning with a unified text-to-text transformer.
\newblock \emph{Journal of machine learning research}.

\bibitem[{Rao et~al.(2024)Rao, Liu, Lin, Ding, Li, Tao, and Zhang}]{rao2024exploring}
Jun Rao, Xuebo Liu, Zepeng Lin, Liang Ding, Jing Li, Dacheng Tao, and Min Zhang. 2024.
\newblock Exploring and enhancing the transfer of distribution in knowledge distillation for autoregressive language models.
\newblock \emph{arXiv preprint arXiv:2409.12512}.

\bibitem[{Rei et~al.(2020)Rei, Stewart, Farinha, and Lavie}]{rei2020comet}
Ricardo Rei, Craig Stewart, Ana~C Farinha, and Alon Lavie. 2020.
\newblock Comet: A neural framework for mt evaluation.
\newblock In \emph{Proceedings of the 2020 Conference on Empirical Methods in Natural Language Processing (EMNLP)}.

\bibitem[{Taori et~al.(2023)Taori, Gulrajani, Zhang, Dubois, Li, Guestrin, Liang, and Hashimoto}]{taori2023stanford}
Rohan Taori, Ishaan Gulrajani, Tianyi Zhang, Yann Dubois, Xuechen Li, Carlos Guestrin, Percy Liang, and Tatsunori~B Hashimoto. 2023.
\newblock Stanford alpaca: An instruction-following llama model.

\bibitem[{Touvron et~al.(2023)Touvron, Martin, Stone, Albert, Almahairi, Babaei, Bashlykov, Batra, Bhargava, Bhosale et~al.}]{touvron2023llama}
Hugo Touvron, Louis Martin, Kevin Stone, Peter Albert, Amjad Almahairi, Yasmine Babaei, Nikolay Bashlykov, Soumya Batra, Prajjwal Bhargava, Shruti Bhosale, et~al. 2023.
\newblock Llama 2: Open foundation and fine-tuned chat models.
\newblock \emph{arXiv preprint}.

\bibitem[{Vaswani(2017)}]{vaswani2017attention}
A~Vaswani. 2017.
\newblock Attention is all you need.
\newblock \emph{Advances in Neural Information Processing Systems}.

\bibitem[{Wang et~al.(2021)Wang, Yan, Meng, and Zhou}]{wang2021selective}
Fusheng Wang, Jianhao Yan, Fandong Meng, and Jie Zhou. 2021.
\newblock Selective knowledge distillation for neural machine translation.
\newblock \emph{arXiv preprint arXiv:2105.12967}.

\bibitem[{Wang et~al.(2023)Wang, Lyu, Ji, Zhang, Yu, Shi, and Tu}]{wang2023document}
Longyue Wang, Chenyang Lyu, Tianbo Ji, Zhirui Zhang, Dian Yu, Shuming Shi, and Zhaopeng Tu. 2023.
\newblock Document-level machine translation with large language models.
\newblock \emph{arXiv preprint arXiv:2304.02210}.

\bibitem[{Wu et~al.(2024)Wu, Tao, Wang, Zhao, and Wong}]{wu2024rethinking}
Taiqiang Wu, Chaofan Tao, Jiahao Wang, Zhe Zhao, and Ngai Wong. 2024.
\newblock Rethinking kullback-leibler divergence in knowledge distillation for large language models.
\newblock \emph{arXiv preprint arXiv:2404.02657}.

\bibitem[{Xu et~al.(2023)Xu, Kim, Sharaf, and Awadalla}]{xu2023paradigm}
Haoran Xu, Young~Jin Kim, Amr Sharaf, and Hany~Hassan Awadalla. 2023.
\newblock A paradigm shift in machine translation: Boosting translation performance of large language models.
\newblock \emph{arXiv preprint arXiv:2309.11674}.

\bibitem[{Zan et~al.(2022{\natexlab{a}})Zan, Ding, Shen, Cao, Liu, and Tao}]{zan2022bridging}
Changtong Zan, Liang Ding, Li~Shen, Yu~Cao, Weifeng Liu, and Dacheng Tao. 2022{\natexlab{a}}.
\newblock Bridging cross-lingual gaps during leveraging the multilingual sequence-to-sequence pretraining for text generation and understanding.
\newblock \emph{arXiv preprint arXiv:2204.07834}.

\bibitem[{Zan et~al.(2022{\natexlab{b}})Zan, Ding, Shen, Cao, Liu, and Tao}]{zan2022complementarity}
Changtong Zan, Liang Ding, Li~Shen, Yu~Cao, Weifeng Liu, and Dacheng Tao. 2022{\natexlab{b}}.
\newblock On the complementarity between pre-training and random-initialization for resource-rich machine translation.
\newblock In \emph{Proceedings of the 29th International Conference on Computational Linguistics}.

\bibitem[{Zan et~al.(2024)Zan, Ding, Shen, Zhen, Liu, and Tao}]{zan2024building}
Changtong Zan, Liang Ding, Li~Shen, Yibing Zhen, Weifeng Liu, and Dacheng Tao. 2024.
\newblock Building accurate translation-tailored llms with language aware instruction tuning.
\newblock \emph{arXiv preprint arXiv:2403.14399}.

\bibitem[{Zhang et~al.(2022)Zhang, Liu, Meng, Chen, Xu, Liu, and Zhou}]{zhang2022conditional}
Songming Zhang, Yijin Liu, Fandong Meng, Yufeng Chen, Jinan Xu, Jian Liu, and Jie Zhou. 2022.
\newblock Conditional bilingual mutual information based adaptive training for neural machine translation.
\newblock \emph{arXiv preprint arXiv:2203.02951}.

\bibitem[{Zheng et~al.(2023)Zheng, Zhong, Ding, Tian, Niu, Li, and Tao}]{zheng2023self}
Haoqi Zheng, Qihuang Zhong, Liang Ding, Zhiliang Tian, Xin Niu, Dongsheng Li, and Dacheng Tao. 2023.
\newblock Self-evolution learning for mixup: Enhance data augmentation on few-shot text classification tasks.
\newblock \emph{arXiv preprint arXiv:2305.13547}.

\bibitem[{Zhong et~al.(2023{\natexlab{a}})Zhong, Ding, Liu, Du, and Tao}]{zhong2023can}
Qihuang Zhong, Liang Ding, Juhua Liu, Bo~Du, and Dacheng Tao. 2023{\natexlab{a}}.
\newblock Can chatgpt understand too? a comparative study on chatgpt and fine-tuned bert.
\newblock \emph{arXiv preprint}.

\bibitem[{Zhong et~al.(2023{\natexlab{b}})Zhong, Ding, Liu, Du, and Tao}]{zhong2023self}
Qihuang Zhong, Liang Ding, Juhua Liu, Bo~Du, and Dacheng Tao. 2023{\natexlab{b}}.
\newblock Self-evolution learning for discriminative language model pretraining.
\newblock \emph{arXiv preprint arXiv:2305.15275}.

\bibitem[{Zhong et~al.(2024)Zhong, Ding, Shen, Liu, Du, and Tao}]{zhong2024revisiting}
Qihuang Zhong, Liang Ding, Li~Shen, Juhua Liu, Bo~Du, and Dacheng Tao. 2024.
\newblock Revisiting knowledge distillation for autoregressive language models.
\newblock In \emph{Proceedings of the 62nd Annual Meeting of the Association for Computational Linguistics (Volume 1: Long Papers)}.

\bibitem[{Zhong et~al.(2022)Zhong, Ding, Zhan, Qiao, Wen, Shen, Liu, Yu, Du, Chen et~al.}]{zhong2022toward}
Qihuang Zhong, Liang Ding, Yibing Zhan, Yu~Qiao, Yonggang Wen, Li~Shen, Juhua Liu, Baosheng Yu, Bo~Du, Yixin Chen, et~al. 2022.
\newblock Toward efficient language model pretraining and downstream adaptation via self-evolution: A case study on superglue.
\newblock \emph{arXiv preprint arXiv:2212.01853}.

\bibitem[{Zhu et~al.(2020)Zhu, Xia, Wu, He, Qin, Zhou, Li, and Liu}]{zhuincorporating}
Jinhua Zhu, Yingce Xia, Lijun Wu, Di~He, Tao Qin, Wengang Zhou, Houqiang Li, and Tieyan Liu. 2020.
\newblock Incorporating bert into neural machine translation.
\newblock In \emph{International Conference on Learning Representations}.

\bibitem[{Zhu et~al.(2023{\natexlab{a}})Zhu, Liu, Dong, Xu, Kong, Chen, Li, and Huang}]{zhu2023multilingual}
Wenhao Zhu, Hongyi Liu, Qingxiu Dong, Jingjing Xu, Lingpeng Kong, Jiajun Chen, Lei Li, and Shujian Huang. 2023{\natexlab{a}}.
\newblock Multilingual machine translation with large language models: Empirical results and analysis.
\newblock \emph{arXiv preprint arXiv:2304.04675}.

\bibitem[{Zhu et~al.(2023{\natexlab{b}})Zhu, Lv, Dong, Yuan, Xu, Huang, Kong, Chen, and Li}]{zhu2023extrapolating}
Wenhao Zhu, Yunzhe Lv, Qingxiu Dong, Fei Yuan, Jingjing Xu, Shujian Huang, Lingpeng Kong, Jiajun Chen, and Lei Li. 2023{\natexlab{b}}.
\newblock Extrapolating large language models to non-english by aligning languages.
\newblock \emph{arXiv preprint arXiv:2308.04948}.

\end{thebibliography}




\end{document}